%% file: manuscript.tex
\documentclass[final]{cvpr}

\usepackage{times}
\usepackage{epsfig}
\usepackage{graphicx}
\usepackage{amsmath}
\usepackage{amssymb}
\usepackage{algorithm}
\usepackage{algpseudocode}
\usepackage{subfigure}
\usepackage{lineno}
\usepackage{booktabs}
\usepackage{color}
\usepackage{paralist}
\usepackage{array}
\usepackage{enumitem}
\usepackage{multirow}

\usepackage[pagebackref=true, breaklinks=true, letterpaper=true, colorlinks, bookmarks=false, citecolor=blue,linkcolor=blue, urlcolor=blue]{hyperref}

\def\cvprPaperID{665} 
\def\confYear{CVPR 2021}

\def\best#1{\textcolor[rgb]{0,0,0}{\textbf{#1}}}

\begin{document}

	\title{Decoupled Dynamic Filter Networks}

	\author{Jingkai Zhou$^{12}$\thanks{Work carried out during the visit of J. Zhou and Z. Pi at UC Merced}~~~Varun 
		Jampani$^3$~~~Zhixiong Pi$^{24}$~~~Qiong Liu$^{1}$\thanks{Corresponding author}~~~Ming-Hsuan Yang$^{235}$\\{\small {$^1$South China University of Technology~~~$^2$University of California at Merced~~~$^3$Google Research}}\\{\small {$^4$Huazhong University of Science and Technology~~~~~~$^5$Yonsei University}}}
	
	\maketitle
	
	\input{0_abstract}
	
	\input{1_introduction}

	\input{2_related_works}
	
	\input{3_approach}

	\input{4_feature_extraction}

	\input{5_upsampling}

	\input{6_conclusion}

	\input{7_acknowledgment}

	{\small
		\bibliographystyle{ieee_fullname}
		\bibliography{manuscript}
	}
	
\end{document}

%% file: 0_abstract.tex
\begin{abstract}
	\vspace{-4mm}
	Convolution is one of the basic building blocks of CNN architectures.
	Despite its common use, standard convolution has two main shortcomings: 
	\emph{Content-agnostic} and \emph{Computation-heavy}. 
	Dynamic filters are content-adaptive, 
	while further increasing the computational \text{overhead}.
	Depth-wise convolution is a lightweight variant, 
	but it usually leads to a drop in CNN performance or requires a larger number of channels.
	In this work, we propose the Decoupled Dynamic Filter 
	(DDF) that can simultaneously tackle both of these shortcomings.
	Inspired by recent advances in attention, DDF decouples 
	a depth-wise dynamic filter into spatial and channel dynamic filters. 
	This decomposition considerably reduces the number of parameters and limits computational costs to the same level as depth-wise convolution.
	Meanwhile, we observe a significant boost in performance when 
	replacing standard convolution with DDF in classification networks.
	\text{ResNet50 / 101} get improved by 1.9$\%$ and 1.3$\%$ on the top-1 accuracy, while their computational costs are reduced by nearly half. 
	Experiments on the detection and joint upsampling networks also demonstrate 
	the superior performance of the DDF upsampling variant (DDF-Up) in comparison with standard convolution and 
	specialized content-adaptive layers. The project page with code is available\footnote{\scriptsize\url{https://thefoxofsky.github.io/project_pages/ddf}}.
	
\end{abstract}

%% file: 1_introduction.tex
\vspace{-4mm}
\section{Introduction}
\vspace{-1mm}
Convolution is a fundamental building block of convolutional neural networks (CNNs) that have
seen tremendous success in several computer vision tasks, such as image classification, semantic segmentation, pose estimation, to name a few.
Thanks to its simple formulation and optimized implementations, 
convolution has become a de facto standard to propagate and integrate features across image pixels.
In this work, we aim to alleviate two of its main shortcomings: 
\textit{Content-agnostic} and \textit{Computation-heavy}.
\begin{figure}[t]
\centering 
\includegraphics[width=.95\linewidth]{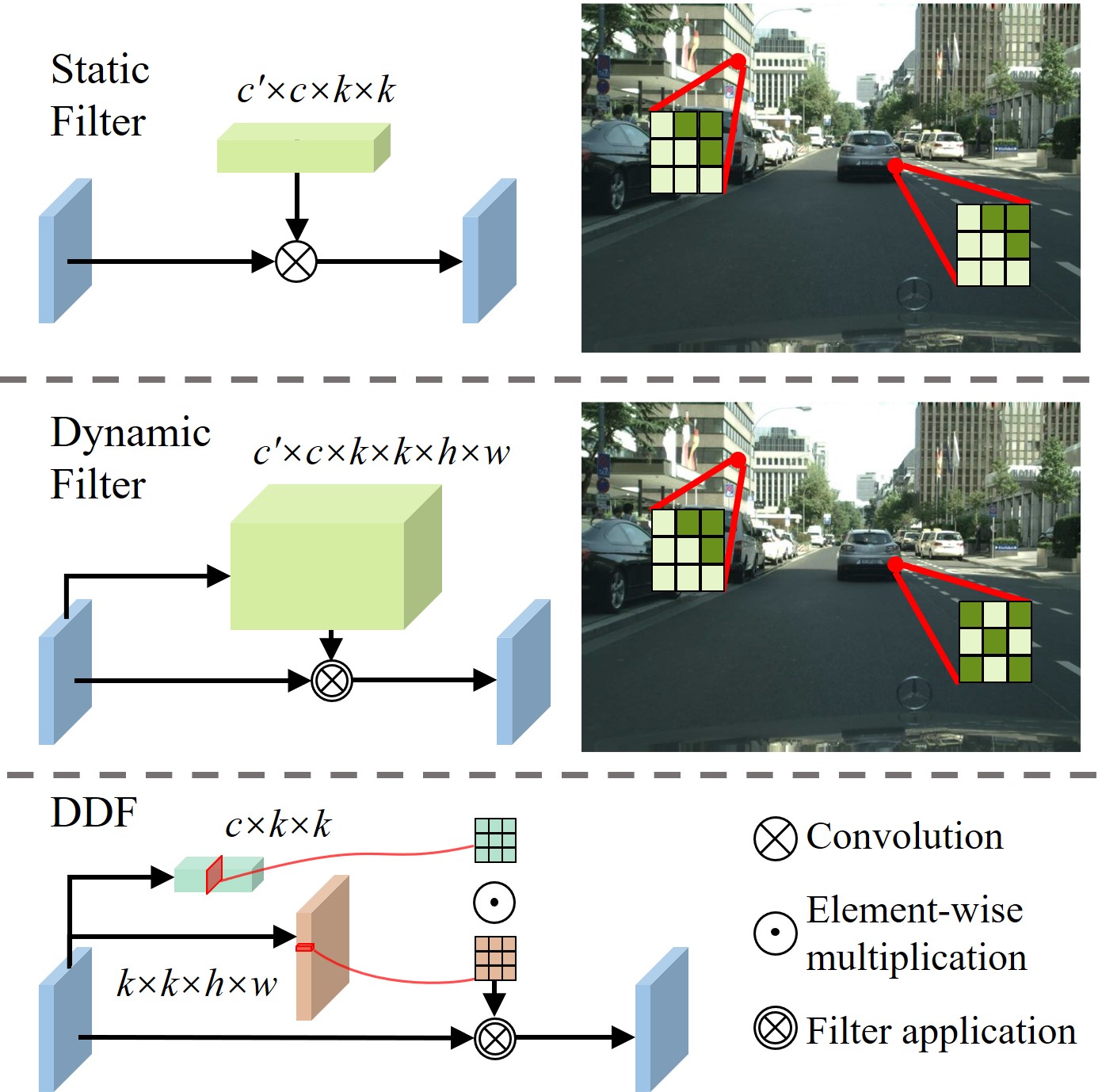} 
\caption{\textbf{Comparison between convolution, the dynamic filter, and DDF.} Top: Convolution shares a static filter among pixels and samples. Medium: The dynamic filter generates one complete filter for each pixel via a separate branch. Bottom: DDF decouples the dynamic filter into spatial and channel ones.} 
\label{fig:dynamic_idea} 
\vspace{-5mm}
\end{figure}

\vspace{1mm}
\noindent \textbf{Content-agnostic.} Spatial-invariance is one of the prominent properties 
of a standard convolution. 
That is, convolution filters are shared across all the pixels in an image.
Consider the sample road scene shown in Figure~\ref{fig:dynamic_idea} (top). 
The convolution filters are shared across different regions such as 
buildings, cars, roads, etc. 
Given the varied nature of contents in a scene, a spatially shared filter 
may not be optimal to capture features across different image regions~\cite{wu2018dynamic,pac}.
In addition, once a CNN is trained, the same convolution filters are used across different images 
(for instance images taken in daylight and at night).
In short, standard convolution filters are content-agnostic and are shared across images and pixels, leading to sub-optimal feature learning.
Several existing works~\cite{dynamic_filter, carafe, pac, adaptive_filter, solov2, cond_seg,jampani2016learning, gadde2016superpixel} 
propose different types of content-adaptive (dynamic) filters for CNNs. 
However, these dynamic filters are either 
compute-intensive~\cite{adaptive_filter, dynamic_filter},
memory-intensive~\cite{pac,jampani2016learning}, or specialized 
processing units~\cite{gadde2016superpixel,carafe, solov2, cond_seg}.
As a result, most of the existing dynamic filters can not completely replace standard convolution
in CNNs and are usually used as a few layers of a CNN~\cite{solov2, cond_seg, pac, jampani2016learning}, or in tiny architecture~\cite{adaptive_filter, dynamic_filter},
or in specific scenarios, like upsampling~\cite{carafe}.

\vspace{1mm}
\noindent \textbf{Computation-heavy.}
Despite the existence of highly-optimized implementations, the computation complexity of 
standard convolution still increases considerably with the enlarge in the filter size or channel number. This poses a significant problem as convolution layers in 
modern CNNs have a large number of channels in the orders of hundreds or even
thousands. Grouped or depth-wise convolutions are commonly used to reduce the computation
complexity. However, these alternatives usually result in CNN performance drops when directly used
as a drop-in replacement to standard convolution. To retain similar performance with
depth-wise or grouped convolutions, we need to considerably increase the number of feature
channels, leading to more memory consumption and access times.

In this work, we propose the Decoupled Dynamic Filter (DDF) that simultaneously addresses both the above-mentioned shortcomings of the standard convolution layer.
The full dynamic filter~\cite{adaptive_filter, dynamic_filter, solov2, cond_seg} uses a separate network branch to predict a complete convolution filter at each pixel. See Figure~\ref{fig:dynamic_idea} (middle) for an illustration. 
We observe that this dynamic filtering is equivalent to applying attention on unfolded input features, as illustrated in Figure~\ref{fig:similarity}.
Inspired by the recent advances in attention mechanisms that apply spatial and channel-wise attention~\cite{bam, cbam}, we propose a new variant of
the dynamic filter where we decouple spatial and channel filters.
In particular, we adopt separate attention-style branches that individually predict spatial and channel dynamic filters,
which are then combined to form a filter at each pixel.
See Figure~\ref{fig:dynamic_idea} (bottom) for an illustration of DDF.
We observe that this decoupling of the dynamic filter is efficient yet effective, making DDF
to have similar computational costs as depth-wise convolution while achieving better performance against existing dynamic filters.
This lightweight nature enables DDF to be directly inserted as a replacement of the standard convolution layer.
Unlike several existing dynamic filtering layers, we can replace \textit{all} $k\times k$ $(k>1)$ convolutions in a CNN with DDF. We also propose a variant of DDF, called DDF-Up, that can be used as a 
specialized upsampling or joint-upsampling layer.

We empirically validate the performance of DDF by drop-in replacing convolution layers in several
classification networks with DDF. Experiments indicate that applying DDF consistently boosts the
performance while reducing computational costs. In addition, we also demonstrate the
superior upsampling performance of DDF-Up in object detection and joint upsampling networks.
In summary, DDF and DDF-Up have the following favorable properties:
\vspace{-2mm}
\begin{itemize}[itemsep=-1mm, leftmargin=*]
    \item \textbf{Content-adaptive.} DDF provides spatially-varying filtering that makes filters adaptive to image contents.
    \item \textbf{Fast runtime.} DDF has similar computational costs as depth-wise convolution, so its inference speed is faster than both standard convolution and dynamic filters. 
    \item \textbf{Smaller memory footprint.} DDF significantly reduces memory consumption of dynamic filters, making it possible to replace all standard convolution layers with DDF.
    \item \textbf{Consistent performance improvements.} Replacing a standard convolution with DDF / DDF-Up results in consistent improvements and achieves the state-of-the-art performance across various networks and tasks.
\end{itemize}

%% file: 2_related_works.tex
\section{Related Work}
\label{sec:related}

\noindent \textbf{Lightweight convolutions.}
Given the prominence of convolutions in CNN architectures, several lightweight variants have been 
proposed for different purposes. 
Dilated convolutions~\cite{chen2017deeplab, dilated} increase the receptive field 
of the filter without increasing parameters
or computation complexity of the standard convolution. Several lightweight mobile networks~\cite{mobile_v1, mobile_v2, mobile_v3} use depth-wise convolutions
instead of standard ones, which separately convolve each channel.
Similarly, grouped convolutions~\cite{alexnet} group input channels and convolve
each group separately resulting in parameter and computation reduction.
However, directly replacing a standard convolution with depth-wise or grouped convolutions usually leads to performance drops. One needs to widen the model to achieve competitive performance with these
lightweight variants of convolution. In contrast, the proposed DDF layer can be directly used as a 
lightweight drop-in replacement to standard convolution layer.

\begin{figure*}[t]
	\centering
	\subfigure[Decoupled Dynamic Filter Operation (DDF Op).]{
		\centering
		\includegraphics[width=.5\linewidth]{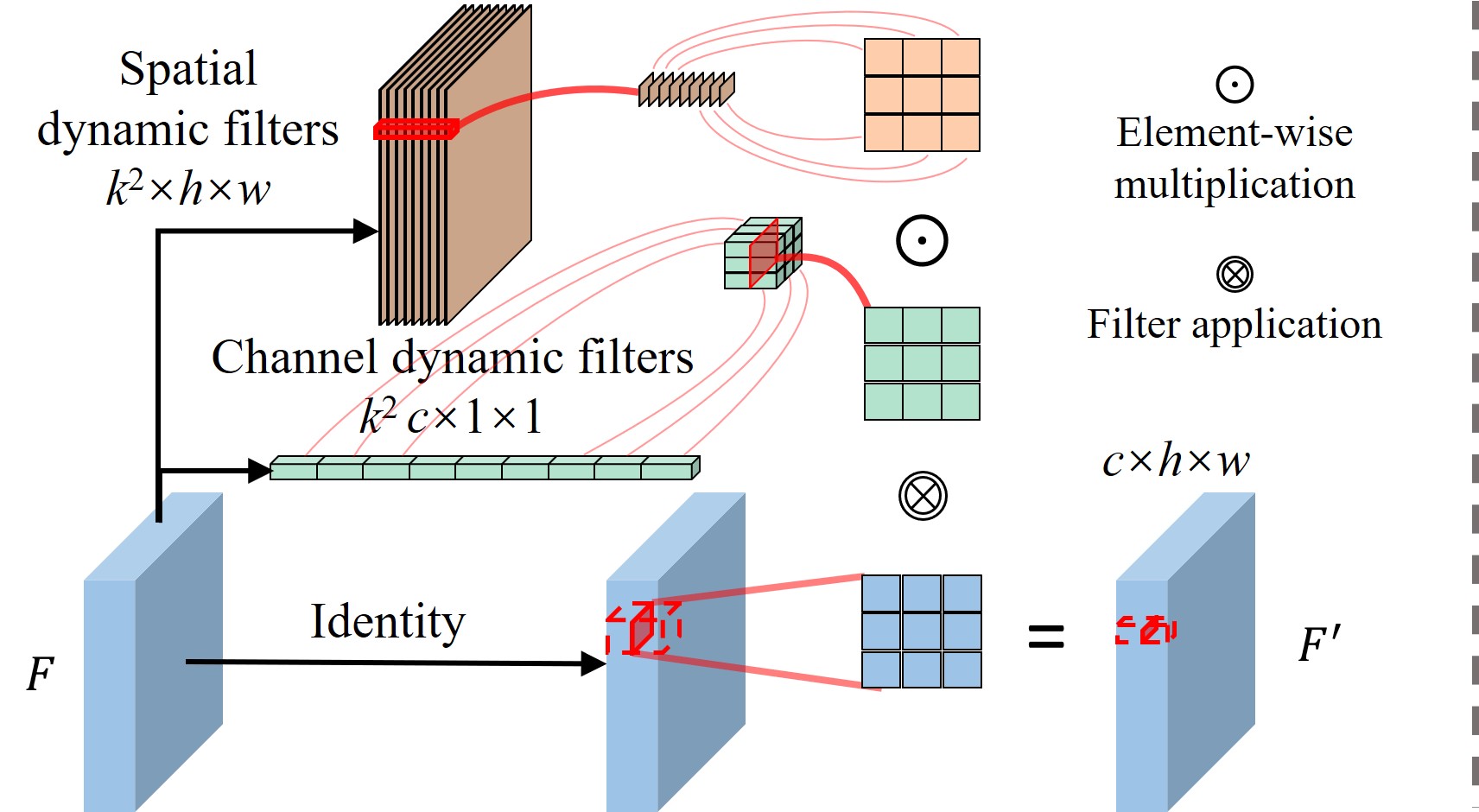}  
		\label{fig:ddf_op}
	}
	\subfigure[Decoupled Dynamic Filter Module (DDF Module).]{
		\centering
		\includegraphics[width=.4\linewidth]{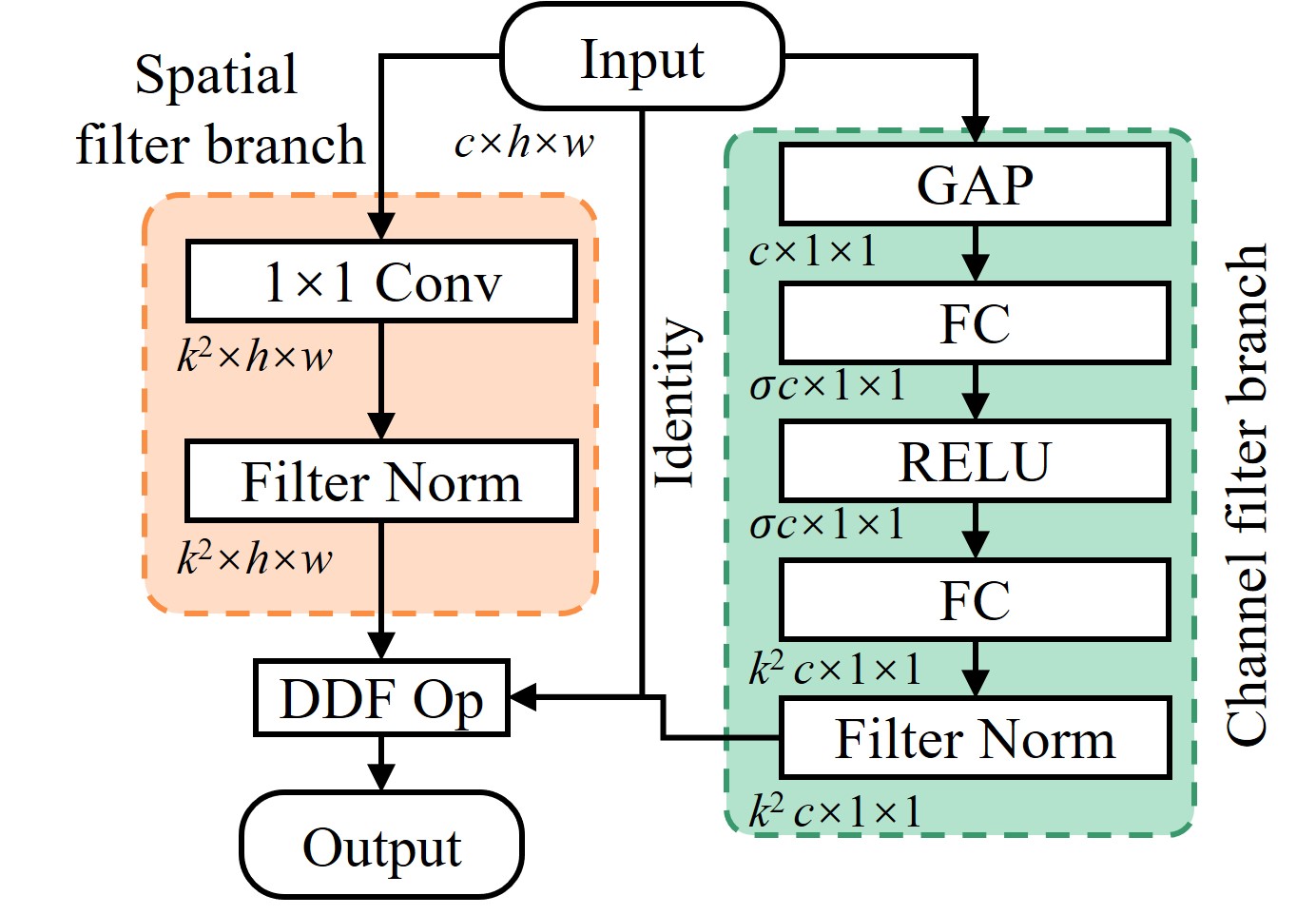}  
		\label{fig:ddf_module}
	}
	\caption{\textbf{Illustration of the DDF operation and the DDF module.} The orange color denotes spatial dynamic filters / branch, and the green color denotes channel dynamic filters / branch. The filter application means applying the convolution operation at a single position. `GAP' means the global average pooling and `FC' denotes the fully connected layer.} 
	\label{fig:ddconv_idea} 
	\vspace{-2mm}
\end{figure*}

\vspace{1mm}
\noindent \textbf{Dynamic filters.} 
For the dynamic filters, the filter neighborhoods and/or filter values 
are dynamically modified or predicted based on the input features. 
Some recent approaches dynamically adjust
the filter neighborhoods by adaptive dilation factors~\cite{scale_adapt_conv},
estimating the neighborhood sampling grid~\cite{deformable_conv}, or adapting the receptive fields~\cite{dau}.
Another kind of dynamic filters, more closely related to our work, 
adjusts or predicts filter values based on input features
~\cite{condconv, dynet, dynamic_conv, dynamic_filter, carafe, pac, adaptive_filter, solov2, cond_seg,jampani2016learning}.
In particular, semi-dynamic filters, such as WeightNet~\cite{weightnet},
CondConv~\cite{condconv}, DyNet~\cite{dynet}, and 
DynamicConv~\cite{dynamic_conv}, predict coefficients to combine several expert filters.
The combined filter is still applied in a convolutional manner (spatially shared).
CARAFE~\cite{carafe} proposes a dynamic layer for upsampling,
where an additional network branch is used
to predict a 2D filter at each pixel. 
However, these channel-wise shared 2D filters cannot encode channel-specific information.
Several full dynamic filters~\cite{dynamic_filter,adaptive_filter,solov2,cond_seg}
use separate network branches to predict a complete filter at each pixel. 
As illustrated in Figure~\ref{fig:ddconv_idea} (middle) and
briefly explained in the Introduction, these dynamic filters can only replace a few convolution layers
or can only be used in small networks due to computational reasons. 
Specifically, adaptive convolutional kernels~\cite{adaptive_filter} are only used in small networks. SOLOv2~\cite{solov2} and CondInst~\cite{cond_seg} employ dynamic filters in the last
few layers of the segmentation model. 
PAC~\cite{pac} uses a fixed Gaussian kernel on adapting features
to modify the standard convolution filter at each pixel, 
which is also impractical for large architectures due to high memory consumption.
The proposed DDF is lightweight even compared with the standard convolution layer and thus can be used
across all the layers even in large networks.

\vspace{1mm}
\noindent \textbf{Attention mechanisms.}
Inspired by the role of attention in human visual perception~\cite{saliency_based, dynamic_repr, control_brain, cbam},
several approaches~\cite{smemvqa, res_att_net, vsgnet, senet, bam, cbam} propose to use attention layers
that dynamically enhance/suppress feature values with predicted attention maps.
SMemVQA~\cite{smemvqa} generates question-guided spatial attention to capture the correspondence between individual words in the question and image regions. The residual attention network~\cite{res_att_net} adopts encoder-decoder branches to model spatial attention and refine features.
VSGNet~\cite{vsgnet} leverages the spatial configuration of human-object pairs to model attention.
Besides spatial attention, SENet~\cite{senet} introduces the squeeze-and-excitation structure to encode channel-wise attention and re-weights the feature channels.
Subsequent methods combine spatial and channel-wise attention. 
BAM~\cite{bam} uses spatial and channel-wise attention in parallel, 
whereas CBAM~\cite{cbam} sequentially applies spatial and channel-wise attention.
In this work, we draw connections between dynamic filters and attention layers. Inspired by spatial
and channel-wise attention, we propose DDF that uses decoupled spatial and channel dynamic filters.

\vspace{-1mm}

%% file: 3_approach.tex
\section{Preliminaries}
\vspace{-1mm}
\label{sec:preliminaries}

\noindent \textbf{Standard convolution.}
Given an input feature representation $F \in \mathbb{R}^{c \times n}$ with $c$ channels and $n$ pixels ($n = h \times w$, $h$ and $w$ are the width and height of the feature map); the standard convolution operation at $i^{th}$ pixel can be written as a linear combination of input features around $i^{th}$ pixel:
\vspace{-2mm}
\begin{equation}
F'_{(.,i)} = \sum_{j\in \Omega(i)}{W[\mathbf{p}_i-\mathbf{p}_j] F_{(.,j)}} + \mathbf{b},
\label{eq:convolution}
\vspace{-3mm}
\end{equation}
where ${F}_{(.,j)} \in \mathbb{R}^c$ denotes the feature vector at $j^{th}$ pixel; $F' \in \mathbb{R}^{c' \times n}$ denotes output feature map with ${F'}_{(.,i)} \in \mathbb{R}^{c'}$ denoting $i^{th}$ pixel output feature vector. $\Omega(i)$ denotes the $k \times k$ convolution window around $i^{th}$ pixel. $W \in \mathbb{R}^{c' \times c \times k \times k}$ is a $k \times k$ convolution filter, $W[\mathbf{p}_i-\mathbf{p}_j] \in \mathbb{R}^{c' \times c}$ is the filter at position offset between $i$ and $j^{th}$ pixels: $[\mathbf{p}_i-\mathbf{p}_j] \in \{(-\frac{(k-1)}{2},-\frac{(k-1)}{2}), (-\frac{(k-1)}{2}, -\frac{(k-1)}{2}+1), ..., (\frac{(k-1)}{2},\frac{(k-1)}{2})\}$ where $\mathbf{p}_{i}$ denotes 2D pixel coordinates. $\mathbf{b} \in \mathbb{R}^{c'}$ denotes the bias vector. In standard convolution, the same filter $W$ is shared across all pixels and filter weights are agnostic to input features.

\vspace{1mm}
\noindent \textbf{Dynamic filters.} 
In contrast to standard convolution, dynamic filters leverage separate network branches to generate the filter at each pixel. The spatially-invariant filter $W$ in Eq.~\ref{eq:convolution} becomes the spatially-varying filter $D_i \in \mathbb{R}^{c' \times c \times k \times k}$ in this case.
The dynamic filters enable learning content-adaptive and flexible feature embeddings. However, predicting such a large number ($nc'ck^2$) of pixel-wise 
filter values requires heavy side-networks, resulting in both compute and memory intensive network architectures. Thus, dynamic filters are usually only employed in either tiny networks~\cite{dynamic_filter, adaptive_filter} or can only replace a few standard convolution layers~\cite{solov2, cond_seg, pac, jampani2016learning} in a CNN.

\section{Decoupled Dynamic Filter}
\vspace{-1mm}

The goal of this work is to design a filtering operation that is \textit{content-adaptive} while being \textit{lighter-weight} than a standard convolution. Realizing both the properties with a single filter is quite challenging. We accomplish this with our Decoupled Dynamic Filter (DDF), where the key technique is to decouple dynamic filters into spatial and channel ones.
More formally, the DDF operation can be written as:
\vspace{-1mm}
\begin{equation}
{F'}_{(r,i)} = \sum_{j\in \Omega(i)}{D^{sp}_{i}[\mathbf{p}_i-\mathbf{p}_j] D^{ch}_{r}[\mathbf{p}_i-\mathbf{p}_j] F_{(r,j)}},
\label{eq:ddconv}
\vspace{-2mm}
\end{equation}
where ${F'}_{(r,i)} \in \mathbb{R}$ denotes the output feature value at the $i^{th}$ pixel and $r^{th}$ channel, ${F}_{(r,j)} \in \mathbb{R}$ denotes the input feature value at the $j^{th}$ pixel and $r^{th}$ channel. $D^{sp} \in \mathbb{R}^{n \times k \times k}$ is the spatial dynamic filter with $D_i^{sp} \in \mathbb{R}^{k \times k}$ denoting the filter at $i^{th}$ pixel. $D^{ch} \in \mathbb{R}^{c \times k \times k}$ is the channel dynamic filter with $D_r^{ch} \in \mathbb{R}^{k \times k}$ denoting the filter at $r^{th}$ channel. 
Figure~\ref{fig:ddf_op} shows the illustration of DDF operation. We predict both channel and spatial dynamic filters from the input feature, using which we perform the above DDF operation (Eq.~\ref{eq:ddconv}) to compute the output feature map. Comparing general dynamic filters (See Section~\ref{sec:preliminaries}) with DDF clearly indicates that DDF reduces the $nc'ck^2$ sized dynamic filter into much smaller $nk^2$ spatial and $ck^2$ channel dynamic filters. In addition, we implement DDF operation in CUDA alleviating any need to save intermediate multiplied filters during network training and inference.

\begin{figure}[t]
	\centering 
	\includegraphics[width=.9\linewidth]{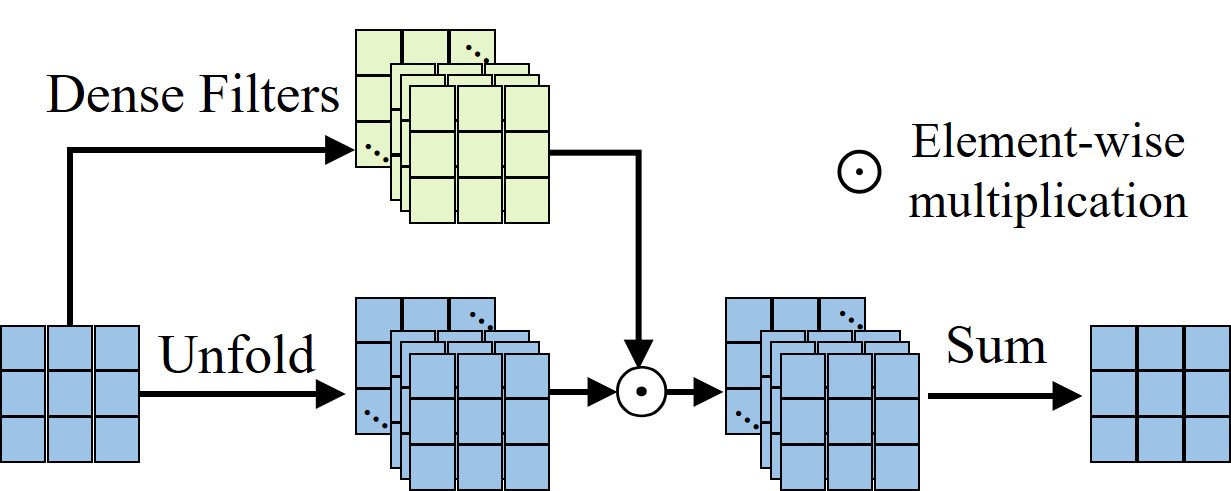} 
	\caption{\textbf{Connection between dynamic filters and attention.} The dynamic filter is similar to applying attention on the unfolded feature.}
	\vspace{-5mm}
	\label{fig:similarity} 
\end{figure}

\vspace{1mm}
\noindent \textbf{DDF module.} 
Based on DDF operation, we carefully design a DDF module that can act as a basic building block in CNNs. For that, we want the filter prediction branches to be lightweight as well in addition to the DDF operation itself. We notice the connection between dynamic filters and attention mechanisms, using which we design attention-style branches to predict spatial and channel filters. Figure~\ref{fig:similarity} illustrates the connection between dynamic filters and attention. Applying dynamic filters on a feature map is equivalent to applying attention on unfolded features. That is, we unfold the $F \in c \times n$ feature map into $F^{u} \in c \times n \times k^2$ feature map where neighboring feature values are unfolded as separate channels. Applying dynamic filters on the original feature map $F$ is the same as re-weighting the unfolded feature map $F^{u}$ using the generated filter tensor as attention.

Following the recent advances in attention literature~\cite{bam, cbam} that propose to use lightweight branches to predict spatial and channel-wise attention, we design two attention-style branches that can generate spatial and channel dynamic filters for DDF. Figure~\ref{fig:ddf_module} illustrates the structure of spatial and channel filter branches in the DDF module. The spatial filter branch only contains one $1 \times 1$ convolution layer. The channel filter branch first applies the global average pooling to aggregate input features, then generates channel dynamic filters via a squeeze-and-excitation structure~\cite{senet}, where the squeeze ratio is denoted as $\sigma \in \mathbb{R}^{+}$.

As generated filter values can be extremely large or small for some input features, directly using them for convolution will make the training unstable. So, we propose to do filter normalization (FN):
\vspace{-2mm}
\begin{equation}
\begin{split}
& D^{sp}_{i} = \alpha^{sp} \frac{\hat{D}^{sp}_i-\mu(\hat{D}^{sp}_i)}{\delta(\hat{D}^{sp}_i)}+\beta^{sp}\\
& D^{ch}_r = \alpha^{ch}_r \frac{\hat{D}^{ch}_r - \mu(\hat{D}^{ch}_r)}{\delta(\hat{D}^{ch}_r)} + \beta^{ch}_r,
\end{split}
\label{filter norm}
\vspace{-2mm}
\end{equation}
where $\hat{D}^{sh}_i, \hat{D}^{ch}_r  \in \mathbb{R}^{k \times k}$ are the generated spatial and channel filters before normalization, $\mu(\cdot)$ and $\delta(\cdot)$ calculate the mean and standard deviation of the filter, $\alpha^{sp}, \alpha^{ch}_r, \beta^{sp}, \beta^{ch}_r$ are the running standard deviation and mean values which are similar to those coefficients in the batch normalization (BN)~\cite{batch_norm}. FN can limit generated filter values into a reasonable range, thereby avoiding the gradient vanishing/exploding during training.

\subsection{Computational Complexity.}
\label{sec:computational_complexity}

Table~\ref{tbl:parameter_comp} shows the parameter, space and time complexity comparisons between standard convolution (Conv), Depth-wise convolution (DwConv), full dynamic filters (DyFilter)~\cite{dynamic_filter, adaptive_filter, solov2, cond_seg}, and our DDF filter. For analysis, we use the same notation as before - $n:$ Number of pixels; $c$: Channel number; $k:$ Filter size (spatial extent); $\sigma:$ Squeeze ratio in DDF channel filter branch. For simplicity, we assume that both input and output features have $c$ channels. We also assume that DyFilter adopts a lightweight filter prediction branch with a single $1 \times 1$ convolution layer.

\vspace{1mm}
\noindent \textbf{Number of parameters.}
The prediction branch of DyFilter takes $c$ channel features as input and produces $c^2k^2$ channel output, where each pixel output corresponds to a complete filter at that pixel. Thus, the DyFilter prediction branch has $c^3k^2$ parameters, which is quite high even for small values of $c$. For DDF, the spatial filter branch predicts filter tensors with $k^2$ channels and thus contain $ck^2$ parameters. The channel filter branch has $\sigma c^2$ parameters for the squeeze layer, and $\sigma c^2k^2$ parameters for the excitation layer. In total, DDF prediction branches contain $ck^2+\sigma c^2(1+k^2)$ parameters, which is far fewer than those for DyFilter. Depending on the values of $\sigma$, $k$, and $c$ (usually set to 0.2, 3, and 256), the number of parameters for the DDF module can be even lower than a standard convolution layer.

\vspace{1mm}
\noindent \textbf{Time complexity.}
The spatial filter generation of DDF needs $2nck^2$ floating-point operations (FLOPs), and the channel filter generation takes $2\sigma c^2(1+k^2)$ FLOPs. The filter combination and application needs $3nck^2$ FLOPs. In total, DDF needs $5nck^2 + 2\sigma c^2(1+k^2)$ FLOPs with time complexity of $O(nck^2 + c^2k^2)$. The term $c^2k^2$ can be ignored since $n>>c,k$. Thus, the time complexity of DDF approximately equals to $O(nck^2)$, which is similar to that of depth-wise convolution and better than a standard convolution with time complexity of $O(nc^2k^2)$.
The time complexity of DyFilter is $O(nc^3k^2)$, with $2nc^3k^2$ FLOPs for filter generation and $2nc^2k^2$ FLOPs for filter application. Thus the time complexity of DyFilter is almost $c^2$ times higher than that of DDF, which is quite significant. Table~\ref{tbl:inference} compares the inference time between four kinds of filters, where we adopt PAC~\cite{pac} as the representative of dynamic filters. 
Refer to the supplementary for more latency comparisons on different input sizes.

\begin{table}[t]
	\caption{\textbf{Comparison of the parameter number and computational costs.} `Params' means the number of parameters, `Time' represents the time complexity, `Space' denotes the space complexity of generated filters.}
	\small
	\centering
	\begin{tabular}{p{0.08\linewidth}p{0.12\linewidth}<{\centering}p{0.1\linewidth}<{\centering}p{0.12\linewidth}<{\centering}p{0.3\linewidth}<{\centering}}
		\toprule
		Filter & Conv & DwConv & DyFilter & DDF \\
		\midrule
		Params & $c^2k^2$   & $ck^2$ & $c^3k^2$ & $ck^2+\sigma c^2(1+k^2)$ \\
		Time & $O(nc^2k^2)$ & $O(nck^2)$ & $O(nc^3k^2)$ & $O(nck^2+c^2k^2)$\\
		Space & -- & -- &  $O(nc^2k^2)$ & $O((n+c)k^2)$ \\
		\bottomrule
	\end{tabular}
	\label{tbl:parameter_comp}
\end{table}

\begin{table}[t]
	\caption{\textbf{Comparison of the inference latency and the max allocated memory.} The size of the input feature is set to $2\times 256 \times 200\times 300$, which is the common size of the P1 layer in FPN~\cite{fpn}. The guidance feature size of PAC is the same as the input one.}
	\small
	\centering
	\begin{tabular}{lcccc}
		\toprule
		Filter & Conv & DwConv & PAC & DDF \\
		\midrule
		Memory & 356.3M & 236.0M & 3406.4M & 245.7M \\
		Latency &  7.5 ms & 1.0 ms & 46.4 ms & 3.0 ms \\
		\bottomrule
	\end{tabular}
	\label{tbl:inference}
	\vspace{-3mm}
\end{table}

\vspace{1mm}
\noindent \textbf{Space/Memory complexity.} Table~\ref{tbl:parameter_comp} also compares the space complexity of generated filters. Standard and depth-wise convolutions do not generate content-adaptive filters. DyFilter generates a complete filter at each pixel with a space complexity of $O(nc^2k^2)$. DDF has a much smaller space complexity of $O((n+c)k^2)$, since it only needs to store 2d spatial filters with $nk^2$ (shared by channels) and channel filters with $ck^2$. See Table~\ref{tbl:inference} for the comparison of the max allocated memory between four kinds of filters.

\vspace{1mm}
In summary, DDF has a time complexity that is similar to depth-wise convolution, which is considerably better than a standard convolution or dynamic filter. Remarkably, despite generating content-adaptive filters, the number of parameters in a DDF module is still smaller than that of a standard convolution layer.
The space complexity of DDF can be hundreds or even thousands of times smaller than full dynamic filters, when $c$ or $n$ are in the orders of hundreds which is quite common in practice.

%% file: 4_feature_extraction.tex
\vspace{-1.8mm}
\section{DDF Networks for Image Classification}
\label{sec:classification}
\vspace{-1.8mm}

\begin{figure}[t]
	\centering 
	\includegraphics[width=.85\linewidth]{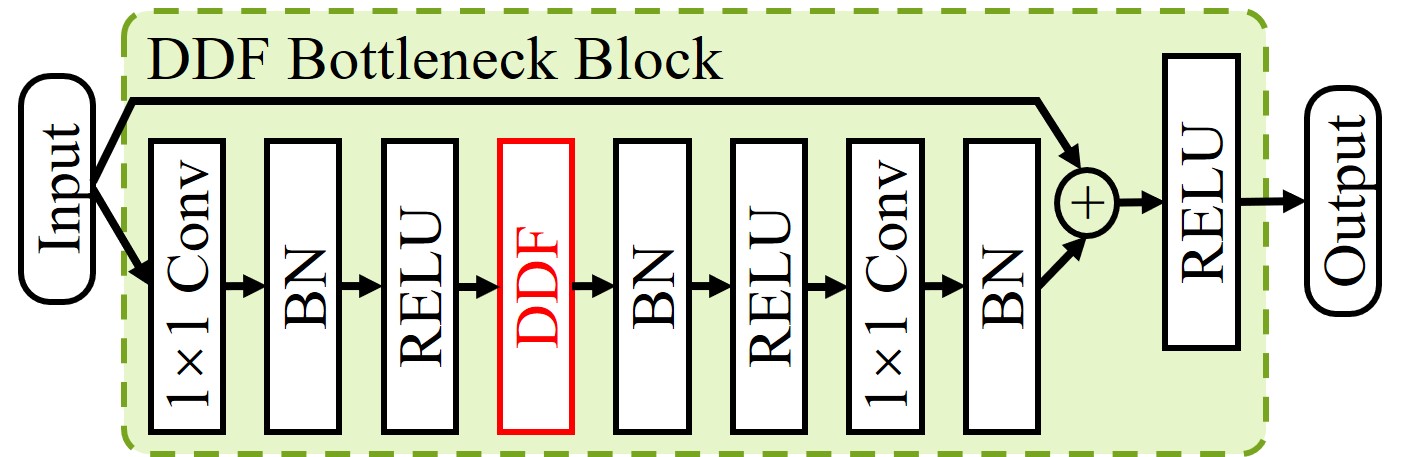} 
	\caption{\textbf{Structure of the DDF bottleneck block.} We replace the $3\times3$ convolution layer with DDF and keep the original hyperparameters, especially using the same number of channels.}
	\label{fig:ddf_bottleneck} 
	\vspace{-3mm}
\end{figure}

Image classification is considered as a fundamental
task in computer vision.
To demonstrate the use of DDFs as basic building blocks in a CNN, we
experiment with the widely used ResNet~\cite{resnet} architecture for
image classification.
ResNets stack multiple basic/bottleneck blocks in which
$3\times3$ convolution layers are adopted for spatial embedding.
We substitute these $3\times3$ convolution layers in \emph{all} 
stacked blocks with DDF. We refer to such a modified
ResNet with DDF as `DDF-ResNet'. Figure~\ref{fig:ddf_bottleneck} illustrates
the use of DDF in a ResNet bottleneck block, we refer to it as
DDF bottleneck block. 

We evaluate DDF-ResNets on the ImageNet dataset~\cite{imagenet} with the Top-1 and Top-5 accuracy. DDF-ResNets are trained using the same training protocol as~\cite{psconv}. In particular, we train models for 120 epochs by the SGD optimizer with the momentum of 0.9 and the weight decay of 1e-4. 
The learning rate is set to 0.1 with batch size 256 and decays to 1e-5 following the cosine schedule. The input image is resized and center-cropped to $224\times224$. 

\vspace{1mm}
\noindent \textbf{Ablation study.}
We comprehensively analyze the effect of different components in a DDF module. 
We choose ResNet50~\cite{resnet} as our base network architecture and experiment
with different modifications to DDF.
Table~\ref{tbl:ablation} shows the results of ablation experiments.
First, we analyze the effect of spatial and channel dynamic filters in DDF with classification
accuracy. Table~\ref{tbl:component} shows there is a significant drop in performance
when we replace convolutions with only spatial dynamic filters.
This is expected as spatial dynamic filters are shared by all channels, thus cannot encode channel-specific information.

By replacing the convolution with the channel dynamic filters, the top-1 accuracy is improved by 1.6\%.
Using the full DDF module, with both spatial and channel dynamic filters, improves the top-1 accuracy
by 1.9\%. These results show the importance of both the spatial and channel dynamic filters in DDF.

Table~\ref{tbl:norm} compares different normalization schemes in a DDF module.
Replacing the proposed filter normalization with a 
standard batch normalization~\cite{batch_norm} or a sigmoid activation leads to
considerable drops in accuracy.
Sigmoid activation individually processes each filter value and may not capture the correlation between them, while batch normalization considers all the filters in a batch, which may weaken the filter dynamics across samples.

We also evaluate DDF under different squeeze ratios $\sigma$, which is used to control the feature channel compression in the channel filter branch. As shown in table~\ref{tbl:hyper}, using higher squeeze ratios will significantly increase the number of parameters, while only bringing marginal performance improvements. Hence, we set the squeeze ratio to 0.2 by default. 
In addition, even the parameter number increases with enlarging the squeeze ratio,
the FLOPs remain low because the computational costs 
of the channel filter branch are minimal, as analyzed in Section~\ref{sec:computational_complexity}.

\begin{table}[t]
	\caption{\textbf{Ablation studies on the ImageNet dataset.} We list the classification performance of
		different DDF-ResNet50 variants, where we use ResNet50 as the base network architecture.}
	\small
	\centering
	\subtable[Effect of spatial and channel filters in DDF.]{
		\begin{tabular}{ccc}
			\toprule
			Spatial & Channel & Top-1 / Top-5 Acc.\\
			\midrule
			\multicolumn{2}{c}{\textit{Base Model}} & 77.2 / 93.5 \\
			$\checkmark$ &  & 74.4 / 92.0 \\
			& $\checkmark$  & 78.7 / 94.2 \\
			$\checkmark$ & $\checkmark$  & \best{79.1} / \best{94.5} \\
			\bottomrule
		\end{tabular}
		\label{tbl:component}}
	
	\centering
	\subtable[Comparison of different normalization schemes.]{
		\begin{tabular}{lccc}
			\toprule
			& Batch-Norm & Sigmoid & \textbf{Filter-Norm}\\
			Top-1 Acc. & 76.0 & 78.2 & \best{79.1}\\
			Top-5 Acc. & 92.0 & 93.8 & \best{94.5}\\
			\bottomrule
		\end{tabular}
		\label{tbl:norm}}
	
	\centering
	\subtable[Comparisons with different squeeze ratios $\sigma$.]{
		\begin{tabular}{lccc}
			\toprule
			$\sigma$  & Params & FLOPs & Top-1 / Top-5 Acc\\
			\midrule
			\textbf{0.2} & \best{16.8M} & \best{2.298B} & 79.1 / \best{94.5}\\
			0.3 & 18.1M & 2.299B & 79.0 / \best{94.5}\\
			0.4 & 19.4M & 2.300B & \best{79.2} / \best{94.5}\\
			\bottomrule
		\end{tabular}
		\label{tbl:hyper}}
	\label{tbl:ablation}
\end{table}

\begin{table}[t]
	\caption{\textbf{Comparison against related filters on the ImageNet dataset.} `--' denotes the unreported value.}
	\small
	\centering
	\begin{tabular}{llccc}
		\toprule
		Arch & Conv Type & Params & FLOPs & Top-1 Acc\\
		\midrule
		\multirow{4}{*}{R18} & \textit{Base Model}~\cite{resnet} & 11.7M & 1.8B & 69.6 \\
		& Adaptive~\cite{adaptive_filter} & 11.1M & -- & 70.2\\
		& DyNet~\cite{dynet} & 16.6M & 0.6B & 69.0\\
		& \textbf{DDF} & \best{7.7M} & \best{0.4B} & \best{70.6}\\
		\midrule
		\multirow{5}{*}{R50} & \textit{Base Model}~\cite{resnet} & 25.6M & 4.1B & 77.2\\
		& DyNet~\cite{dynet} & -- & \best{1.1B} & 76.3\\
		& CondConv~\cite{condconv} & 104.8M & 4.2B & 78.6\\
		& DwCondConv~\cite{condconv} & 14.5M & 2.3B & 78.3\\
		& DwWeightNet~\cite{weightnet} & \best{14.4M} & 2.3B & 78.0\\
		& \textbf{DDF} & 16.8M & 2.3B & \best{79.1}\\
		\bottomrule
	\end{tabular}
	\label{tbl:dynamic_filters}
	\vspace{-2mm}
\end{table}

\begin{table}[t]
	\caption{\textbf{Comparison with state-of-the-art variants of ResNet50 and ResNet101 on the ImageNet dataset.} Variants include attention mechanisms: SE, BAM, CBAM, AA; and block modifications: ResNeXt, Res2Net, and our DDF. Besides official results from the respective work, we list re-trained results (in brackets) under the same training protocol (that we use) as in~\cite{psconv}.}
	\small
	\begin{center}
		\begin{tabular}{lccc}
			\toprule
			Method  & Params & FLOPs & Top-1 Acc\\
			\midrule
			\textit{ResNet50 (base)}~\cite{resnet} & 25.6M & 4.1B & 76.0 (77.2)\\
			SE-ResNet50~\cite{senet} & 28.1M & 4.1B & 77.6 (77.8)\\
			BAM-ResNet50~\cite{bam} & 25.9M & 4.2B & 76.0\\
			CBAM-ResNet50~\cite{cbam} & 28.1M & 4.1B & 77.3 \\
			AA-ResNet50~\cite{aa} & 25.8M & 4.2B & 77.7\\ 
			ResNeXt50 (32$\times$4d)~\cite{resnext} & 25.0M & 4.3B & 77.8 (78.2)\\
			Res2Net50 (14w-8s)~\cite{res2net} & 25.7M & 4.2B & 78.0\\
			\textbf{DDF-ResNet50} & \best{16.8M} & \best{2.3B} & \best{79.1}\\
			\midrule
			\textit{ResNet101 (base)}~\cite{resnet} & 44.5M & 7.8B & 77.6 (78.9) \\
			SE-ResNet101~\cite{senet} & 49.3M & 7.8B & 78.3 (79.3)\\
			BAM-ResNet101~\cite{cbam} & 44.9M & 7.9B & 77.6\\
			CBAM-ResNet101~\cite{cbam} & 49.3M & 7.8B & 78.5\\
			AA-ResNet101~\cite{aa} & 45.4M & 8.1B & 78.7\\ 
			ResNeXt101 (32$\times$4d)~\cite{resnext} & 44.2M & 8.0B & 78.8 (79.5)\\
			Res2Net101 (26w-4s)~\cite{res2net} & 45.2M & 8.1B & 79.2\\
			\textbf{DDF-ResNet101} & \best{28.1M} & \best{4.1B} & \best{80.2}\\
			\bottomrule
		\end{tabular}
	\end{center}
	\label{tbl:variants}
	\vspace{-6mm}
\end{table}

\vspace{1mm}
\noindent \textbf{Comparisons with other dynamic filters.}
Next, we compare the use of DDF with respect to some existing dynamic filters using different ResNet
base architectures: ResNet18 (R18) and ResNet50 (R50).
Table~\ref{tbl:dynamic_filters} shows the parameters, FLOPs, and accuracy comparisons.
Specifically, we compare DDF with 
adaptive convolutional kernel~\cite{adaptive_filter} (Adaptive), 
DyNet~\cite{dynet}, Conditionally parameterized convolutions (CondConv/DwCondConv)~\cite{condconv},
and depth-wise WeightNet (DwWeightNet)~\cite{weightnet}.
The `Adaptive'~\cite{adaptive_filter} can only be used in R18 due to its large running memory consumption.
Results show that using DDF consistently boosts the performance of base models, while also significantly reducing the number of parameters and FLOPs.
It is worth noting that, DwWeightNet has worse performance than DDF, and even inferior to the channel-only DDF in Table~\ref{tbl:component}, although it has a similar design as the channel-only DDF. This is due to the 
use of sigmoid activation during the filter generation in DwWeightNet

(more analysis in the supplementary).

\vspace{1mm}
\noindent \textbf{Comparisons with state-of-the-art ResNet variants.}
We also compare DDF-ResNets with other state-of-the-art variants of ResNet50 and ResNet101
architectures in Table~\ref{tbl:variants}. Specifically, we compare with 
attention mechanisms of SE~\cite{senet}, BAM~\cite{bam}, CBAM~\cite{cbam} and AA~\cite{aa};
and also block modifications of ResNeXt~\cite{resnext} and Res2Net~\cite{res2net}.
Results clearly show that DDF-ResNets achieve the best performance while also having
the lowest number of parameters and FLOPs. DDF-ResNet50 can be further improved by tricks in training and evaluation, and can achieve 81.3\% top-1 accuracy. Refer to the supplementary for more details.

Recently, neural architecture search (NAS) methods~\cite{mnasnet, darts} can obtain architectures with outstanding speed/accuracy trade-off. 
The proposed DDF module can also contribute to the search space of NAS methods as a new fundamental building block.

%% file: 5_upsampling.tex
\vspace{-1mm}
\section{DDF as Upsampling Module}
\vspace{-1mm}
\label{sec:upsampling}

\begin{figure}[t]
	\centering 
	\includegraphics[width=.85\linewidth]{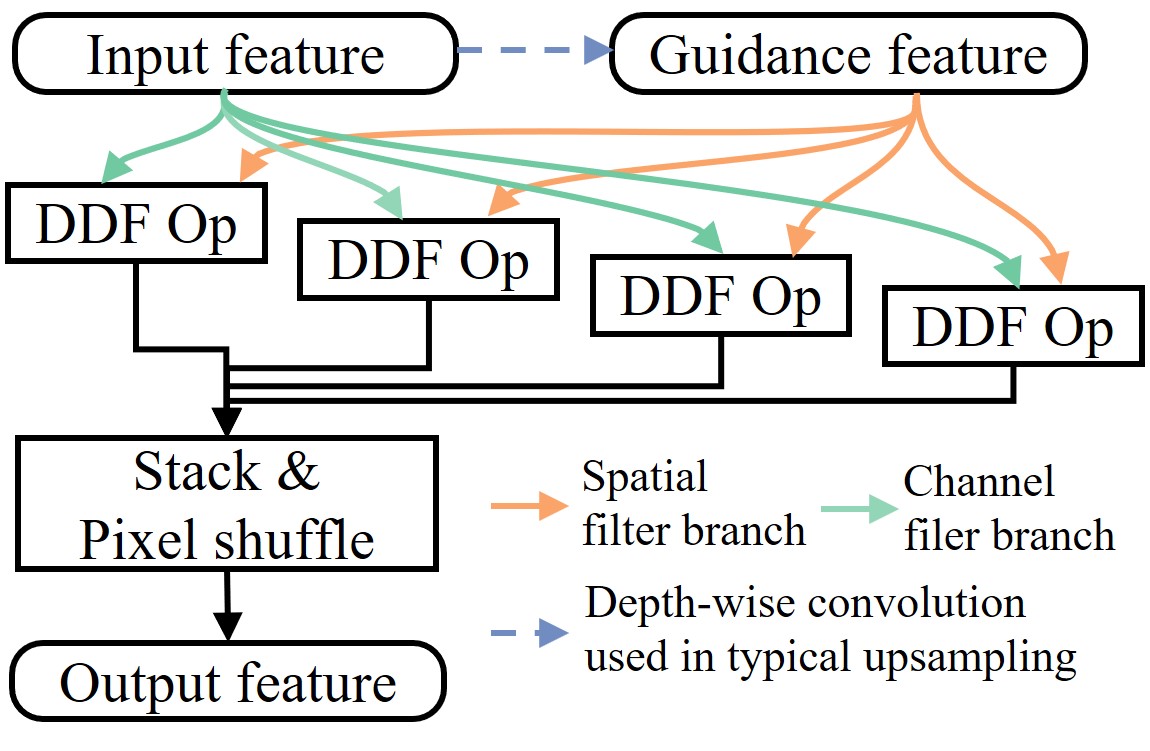} 
	\caption{\textbf{Structure of the DDF-Up module.} When the upsampling scale factor is set to 2, the DDF-Up module contains 4 branches. For typical upsampling, the guidance feature is predicted from input features via a depth-wise convolution layer.}
	\label{fig:ddconv_up} 
	\vspace{2mm}
\end{figure}

\begin{figure}[t]
	\centering 
	\subfigure[FPN with DDF-Up.]{
		\centering
		\includegraphics[width=.9\linewidth]{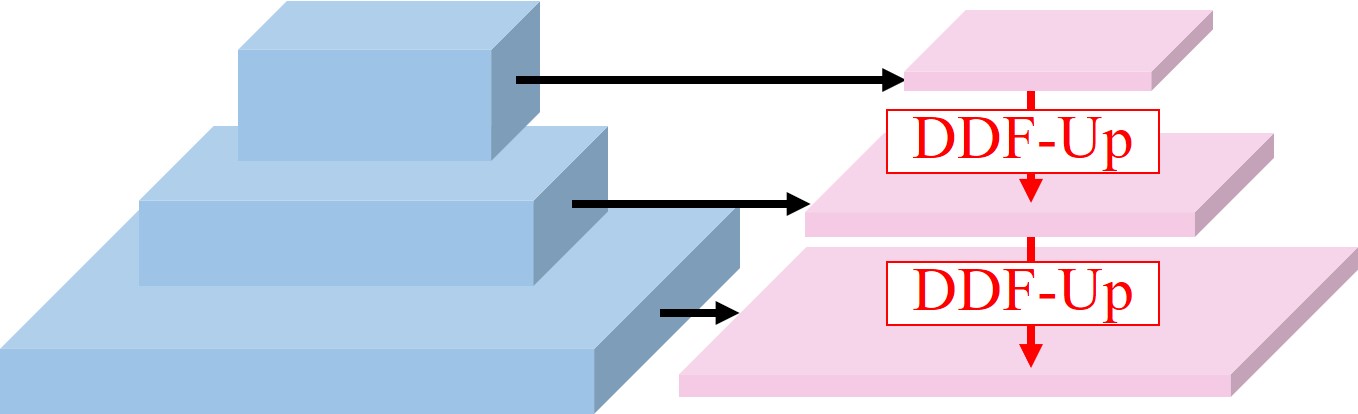}  
		\label{fig:ddf_fpn}
	}
	\subfigure[Joint upsampling with DDF-Up.]{
		\centering
		\includegraphics[width=.9\linewidth]{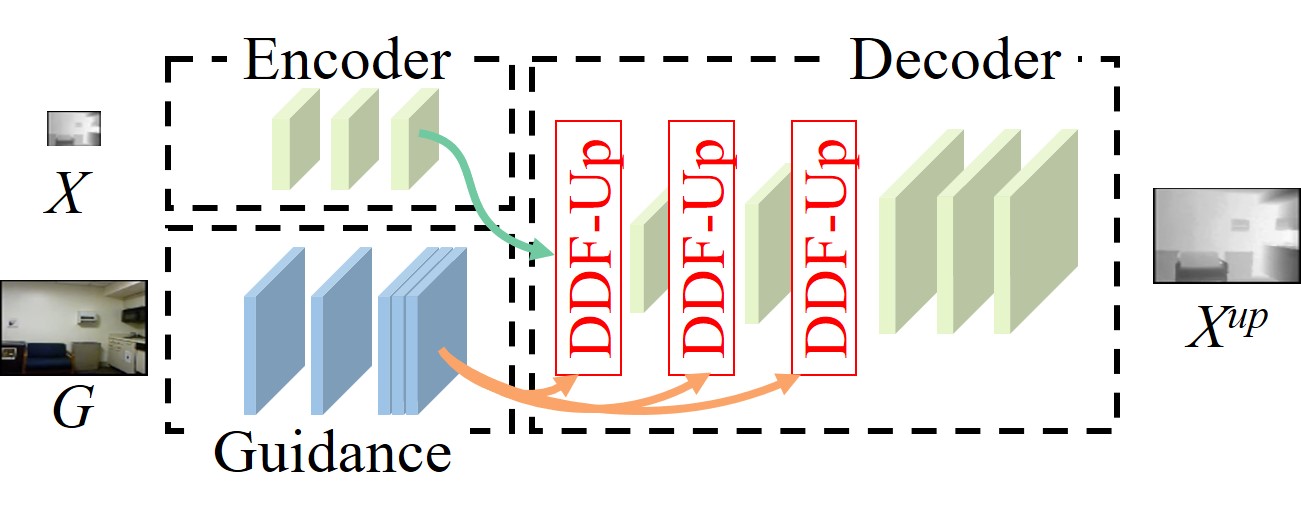}  
		\label{fig:ddf_joint}
	}
	\caption{\textbf{Applications of the DDF-Up module.} DDF-Up can be seamlessly embedded into the top-down 
		upsampling path in the FPN~\cite{fpn} network for object detection and the decoder part of a joint upsampling architecture.}
	\vspace{-2mm}
\end{figure}

An advantage of dynamic filters compared with standard convolution is that one could predict dynamic filters
from \textit{guidance} features instead of input features. Following this, we propose an extension 
of the DDF module, where spatial dynamic filters are predicted 
using separate guidance features instead of input features.
Such joint filtering with input and guidance features is useful for several tasks such as joint image
upsampling~\cite{pac, djf, djf+}, cross-modal image enhancement~\cite{crossmodal1, crossmodal2, crossmodal3}, texture removal~\cite{texture} to name a few.
Figure~\ref{fig:ddconv_up} illustrates the DDF Upsampling (DDF-Up) module, 
where the number of DDF operations used
is set to $x^2$ for the upsampling factor $x$ (e.g., 4 DDF operations when the upsampling factor is 2).
We stack and pixel-shuffle~\cite{pixel_shuffle} the resulting features from the DDF operations to form
output features. For typical upsampling (without guidance), we use the same structure with a slight
modification. We compute guidance features from input ones using a depth-wise convolution layer.
DDF-Up can be seamlessly integrated into several 
existing CNNs, where typical/joint upsampling operators are needed. 
Here we present two applications in object detection and joint depth upsampling tasks.

\begin{table}[t]
	\caption{\textbf{Comparison of different upsampling modules in FPN~\cite{fpn} on the COCO minival split.} We show FLOPs (for upsampling modules) and mAp scores on small (mAp$_{S}$), medium (mAp$_{M}$), large (mAp$_{L}$), and all-scale (mAp) objects.}
	\small
	\begin{center}
		\begin{tabular}{p{0.219\linewidth}p{0.12\linewidth}<{\centering}p{0.08\linewidth}<{\centering}p{0.08\linewidth}<{\centering}p{0.08\linewidth}<{\centering}p{0.08\linewidth}<{\centering}}
			\toprule
			Method & FLOPs & mAp$_{S}$ & mAp$_{M}$ & mAp$_{L}$  & mAp \\
			\midrule
			\textit{Nearest (base)} & \best{0.00B}& 21.2 & 41.0 & 48.1 & 37.4\\
			Bilinear & 0.02B & 22.1 & 41.2 & 48.4 & 37.6\\
			Deconv~\cite{deconvolution} & 12.57B & 21.0 & 41.1 & 48.5 & 37.3\\
			P.S.~\cite{pixel_shuffle} & 50.18B & 21.4 & 41.5 & 48.6 & 37.7\\
			CARAFE~\cite{carafe} & 2.14B & \best{22.6} & 42.0 & 49.8 & 38.5\\
			\textbf{DDF-Up} & 0.58B & 22.1 & \best{42.4}& \best{49.9} & \best{38.6}\\
			\bottomrule
		\end{tabular}
	\end{center}
	\label{tbl:detection}
	\vspace{-3mm}
\end{table}

\vspace{1mm}
\noindent \textbf{Object detection with DDF-Up.}
Detecting objects in an image is one of the core dense prediction tasks in computer vision.
We adopt FasterRCNN~\cite{faster_rcnn} with the 
Feature Pyramid Network (FPN)~\cite{fpn} as our
base detection architecture and embed DDF-Up modules into FPN.
FPN is an effective U-net shaped feature fusion module, where the decoder pathway
upsamples high-level features while combining low-level ones.
As illustrated in Figure~\ref{fig:ddf_fpn}, we replace the nearest-neighbor
upsampling modules in FPN with our DDF-Up modules.

\begin{figure*}
	\centering
	\begin{tabular}{c@{\hspace{1mm}\vspace{.1mm}}c@{\hspace{1mm}\vspace{.1mm}}c@{\hspace{1mm}\vspace{.1mm}}c@{\hspace{1mm}\vspace{.1mm}}c@{\hspace{1mm}\vspace{.1mm}}c}
		\begin{minipage}{0.15\linewidth}
			\includegraphics[width=1.0\linewidth]{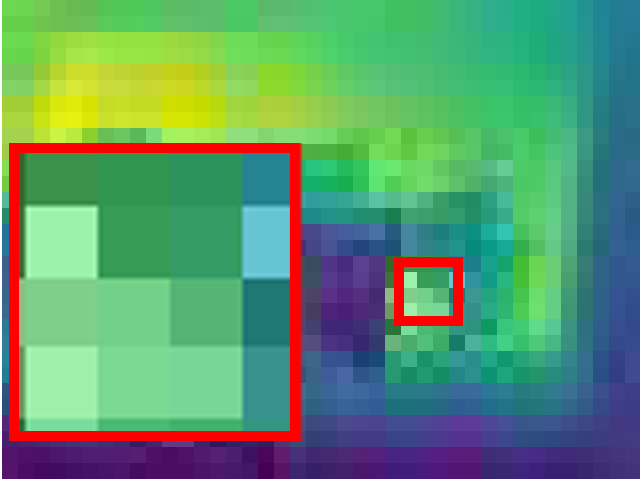}
		\end{minipage} &
		\begin{minipage}{0.15\linewidth}
			\includegraphics[width=1.0\linewidth]{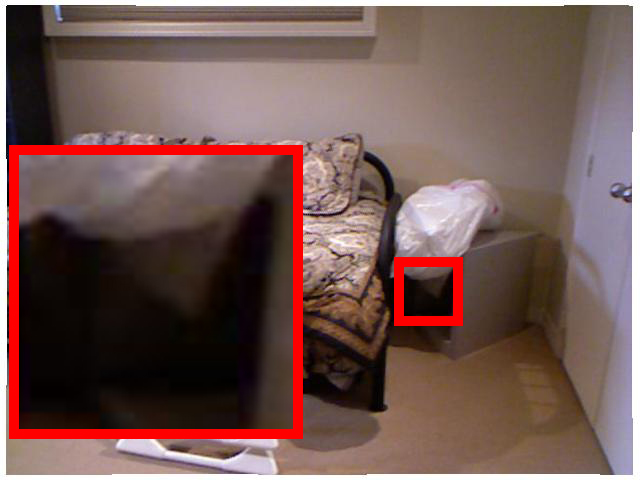}
		\end{minipage} &
		\begin{minipage}{0.15\linewidth}
			\includegraphics[width=1.0\linewidth]{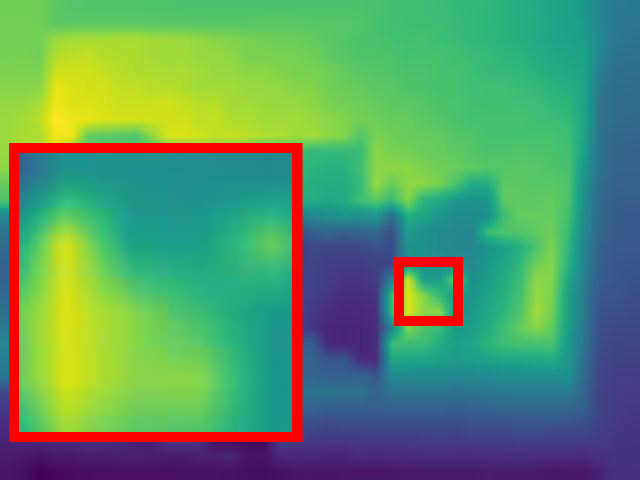}
		\end{minipage} &
		\begin{minipage}{0.15\linewidth}
			\includegraphics[width=1.0\linewidth]{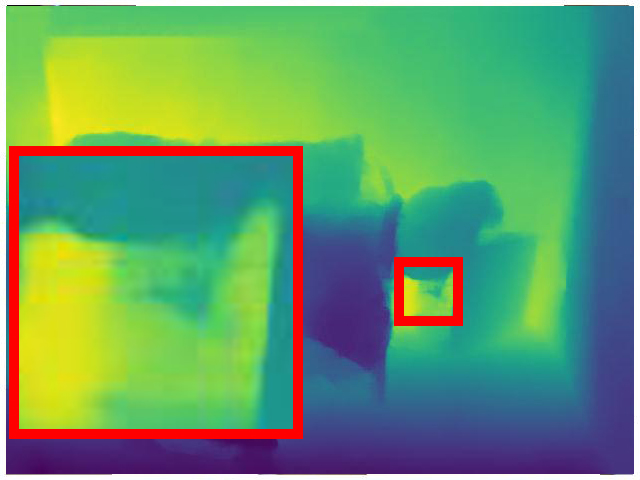}
		\end{minipage} &
		\begin{minipage}{0.15\linewidth}
			\includegraphics[width=1.0\linewidth]{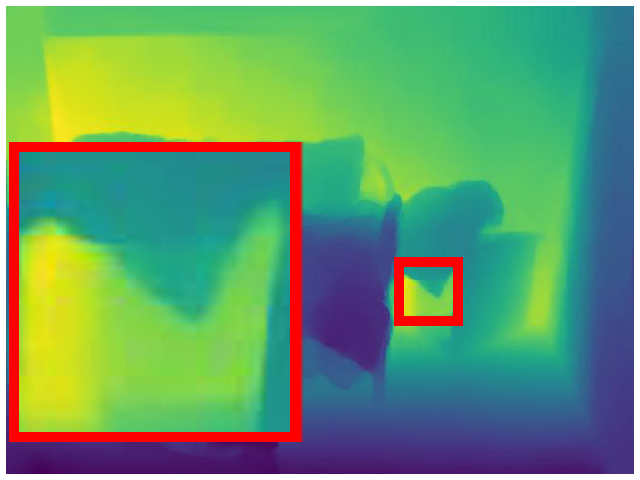}
		\end{minipage} &
		\begin{minipage}{0.15\linewidth}
			\includegraphics[width=1.0\linewidth]{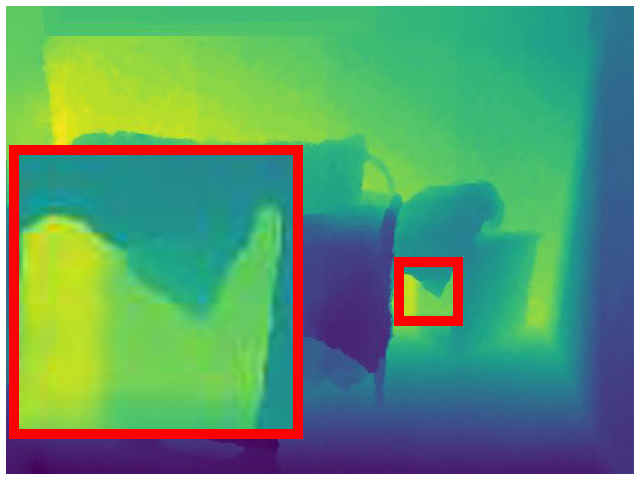}
		\end{minipage} \\
		\begin{minipage}{0.15\linewidth}
			\includegraphics[width=1.0\linewidth]{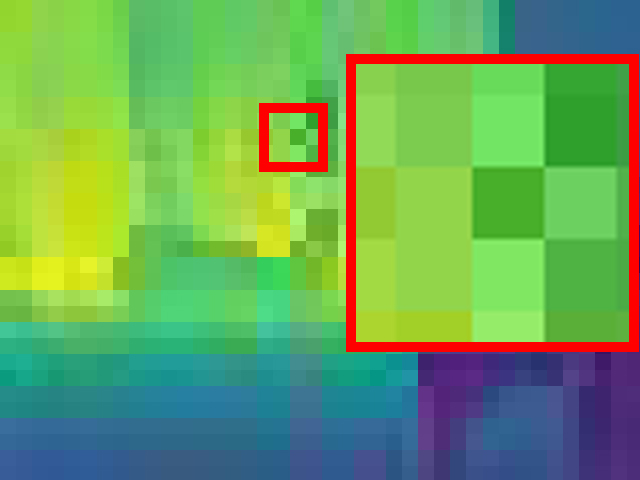}
		\end{minipage} &
		\begin{minipage}{0.15\linewidth}
			\includegraphics[width=1.0\linewidth]{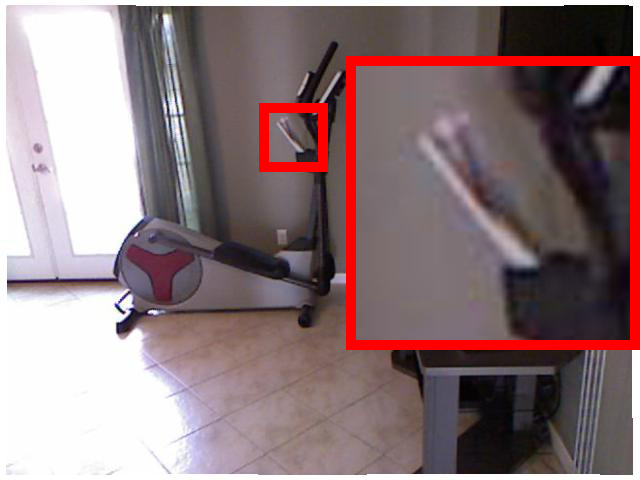}
		\end{minipage} &
		\begin{minipage}{0.15\linewidth}
			\includegraphics[width=1.0\linewidth]{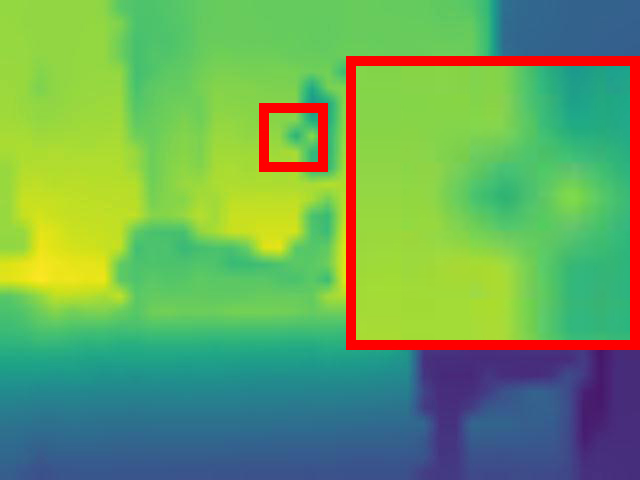}
		\end{minipage} &
		\begin{minipage}{0.15\linewidth}
			\includegraphics[width=1.0\linewidth]{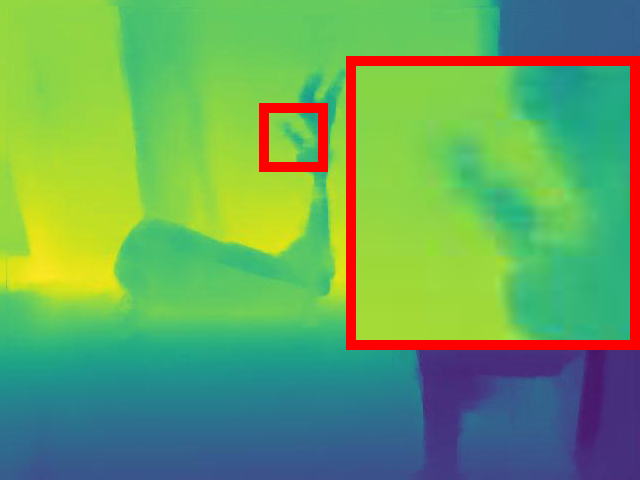}
		\end{minipage} &
		\begin{minipage}{0.15\linewidth}
			\includegraphics[width=1.0\linewidth]{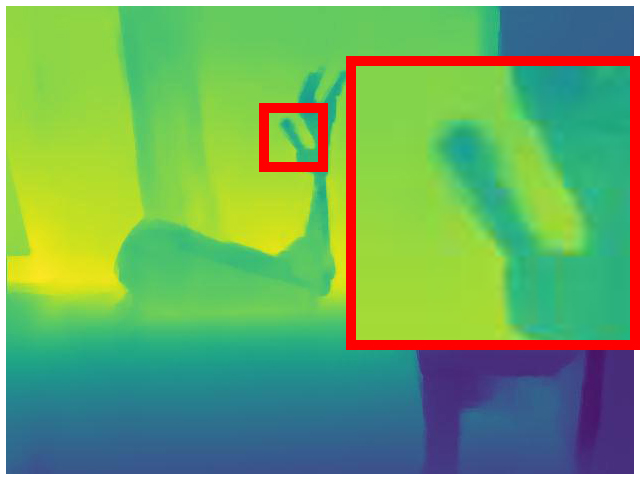}
		\end{minipage} &
		\begin{minipage}{0.15\linewidth}
			\includegraphics[width=1.0\linewidth]{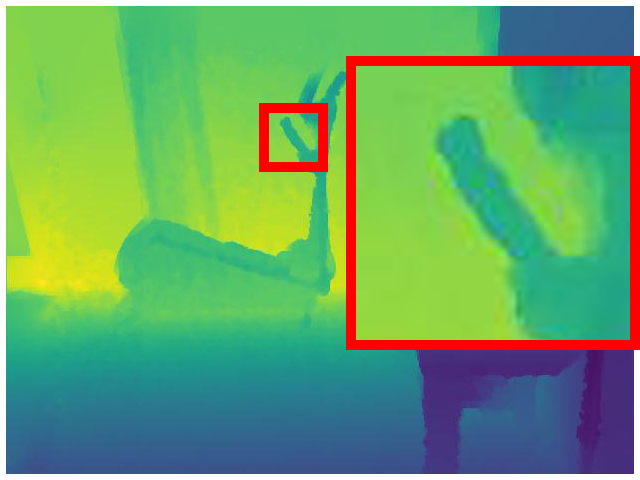}
		\end{minipage} \\
		\small{Input} & \small{Guidance} & \small{Bilinear} & \small{PAC-Net~\cite{pac}} & \small{\textbf{DDF-Up-Net (Ours)}} & \small{GT}
	\end{tabular}
	\caption{\textbf{16$\times$ joint depth sampling results on sample images.} DDF-Up-Net
		recovers more depth details compared with PAC-Net~\cite{pac} and other techniques.}
	\label{fig:vis}
\end{figure*}

We analyze the effectiveness of DDF-Up modules with experiments on the COCO detection
benchmark~\cite{coco} which contains 115K training and 5K validation images.
We report standard COCO~\cite{coco} metrics for small (mAp$_{S}$), medium (mAp$_{M}$), large (mAp$_{L}$), and all-scale (mAp) objects on the
minival split. We implement our models based on the MMDetection~\cite{mmdet} toolbox
and train them using the standard training protocol therein. 
Specifically, we train different models for 12 epochs using the SGD optimizer with a momentum of 0.1
and the weight decay of 1e-4. We use a batch size of 16 and set the learning rate to 0.2 which
decays by the factor of 0.1 at 8 and 11$^{th}$ epochs.
We resize the shorter side of the input image to 800 pixels, while keeping the longer side no larger than 1333 pixels.

We compare DDF-Up with the generic nearest-neighbor (Nearest) and bilinear (Bilinear) interpolations, as well as learnable Deconvolution (Deconv)~\cite{deconvolution}, Pixel Shuffle (P.S.)~\cite{pixel_shuffle}, and CARAFE~\cite{carafe} upsampling modules. 
Table~\ref{tbl:detection} exhibits the comparison results. FPN with DDF-Up yields a 1.2\% mAp improvement over the baseline which adopts the nearest-neighbor interpolation. DDF-Up also brings obvious improvements against static-filtering upsampler (like Deconv and P.S.), and is on par with the recent dynamic upsampling technique CARAFE while utilizing only one-third of FLOPs as CARAFE.

\vspace{1mm}
\noindent \textbf{Joint depth upsampling with DDF-Up.}
We analyze the use of DDF-Up as a joint upsampling module by integrating it into a
joint depth upsampling network. Here, the task is to upsample a low-resolution depth map given a
higher-resolution RGB image as guidance.
This experiment allows to compare DDF-Up with content-adaptive filtering techniques such as 
Pixel-Adaptive Convolution (PAC)~\cite{pac} which is a current state-of-the-art for this task.
We use a similar network architecture to PAC-Net~\cite{pac}, where we employ our DDF-Up modules
instead of PAC joint upsampling modules. We call the resulting network `DDF-Up-Net'.
Figure~\ref{fig:ddf_joint} illustrates DDF-Up-Net where we first encode low-resolution
input features from the given depth map ($X$) and high-resolution guidance features ($G$) from RGB images. Then, we employ DDF-Up in the decoder to joint upsample depth
features with guidance features and obtain high-resolution depth output ($X^{up}$).
Each DDF-Up module does $2\times$ upsampling, we sequentially use $k$ DDF-Up modules
when the upsampling factor is $2^k$.

We conduct experiments on the NYU depth V2 dataset~\cite{nyu} which has 1449 RGB-depth pairs.
Following PAC-Net~\cite{pac}, we use the nearest-neighbor downsampling to generate low-resolution inputs from the ground-truth (GT) depth maps. We split the first 1000 samples for training and the rest for testing.
We train DDF-Up-Net for 1500 epochs using the Adam optimizer~\cite{adam}.
We use a batch size of 8 and set the learning rate to 1e-4 which decays by the factor 0.1 at 1000 and 1350$^{th}$ epochs. During training, the input images are resized and random-cropped to $256\times256$.
Table~\ref{tbl:joint_up} reports Root Mean Square Error (RMSE) scores of different techniques for
three upsampling factors, \ie, $4\times$, $8\times$, and $16\times$.
DDF-Up-Net performs better than state-of-the-art techniques across all the upsampling factors.
It surpasses the standard CNN techniques like DJF~\cite{djf} and DJF+~\cite{djf+} by a large
margin. It also improves over dynamic-filtering PAC~\cite{pac} while reducing computational costs by an order of magnitude.
See Table~\ref{tbl:inference} for the cost comparison between PAC and DDF-Up.
We visualize sampled 16$\times$ upsampling results in Figure~\ref{fig:vis}, where we can see that DDF-Up-Net recovers more details compared to PAC-Net and other techniques.

\begin{table}[t]
	\caption{\textbf{Joint depth upsampling results on the NYU Depth V2 dataset.} We exhibit RMSE results (in the order of $10^{-2}$, lower is better) of different techniques and different upsampling factors.}
	\small
	\begin{center}
		\begin{tabular}{lccc}
			\toprule
			Method & 4$\times$ & 8$\times$ & 16$\times$ \\
			\midrule
			Bicubic & 8.16 & 14.22 & 22.32\\
			MRF (32$\times$4d) & 7.84 & 13.98 & 22.20 \\
			GF~\cite{gf} & 7.32 & 13.62 & 22.03\\
			Ham \etal~\cite{ham} & 5.27 & 12.31 & 19.24\\
			FBS~\cite{fbs} & 4.29 & 8.94 & 14.59\\
			JBU~\cite{jbu} & 4.07 & 8.29 & 13.35\\
			DMSG~\cite{dmsg} & 3.78 & 6.37 & 11.16\\
			DJF~\cite{djf} & 3.54 & 6.20 & 10.21 \\
			DJF+~\cite{djf+} & 3.38 & 5.86 & 10.11 \\
			PAC-Net~\cite{pac} & 2.39 & 4.59 & 8.09\\
			\textbf{DDF-Up-Net} & \best{2.16} & \best{4.40} & \best{7.72} \\
			\bottomrule
		\end{tabular}
	\end{center}
	\label{tbl:joint_up}
	\vspace{-5mm}
\end{table}

%% file: 6_conclusion.tex
\vspace{-1mm}
\section{Conclusion}
\vspace{0mm}

In this work, we propose a lightweight content-adaptive filtering technique called DDF, where
our key strategy is to predict decoupled spatial and channel dynamic filters.
We show that DDF can seamlessly replace standard convolution layers, 
consistently improving the performance of ResNets while also reducing model parameters and
computational costs. In addition, we propose an upsampling variant called DDF-Up, which boosts performance as both a general upsampling module in detection and a joint upsampling module in joint depth upsampling. DDF-Up also is more computationally efficient compared with specialized content-adaptive layers.
Overall, DDF has rich representative capabilities as a content-adaptive filter while also being computationally cheaper than a standard convolution, making it highly practical to use in modern CNNs.

%% file: 7_acknowledgment.tex
\vspace{-3mm}
\section{Acknowledgement}
\vspace{-1mm}
This work is supported in part by the National Natural Science Foundation of China (No.61976094). M.-H. Yang is supported in part by NSF CAREER 1149783.

%% file: manuscript.bbl
\begin{thebibliography}{10}\itemsep=-1pt

\bibitem{fbs}
Jonathan~T Barron and Ben Poole.
\newblock The fast bilateral solver.
\newblock In {\em ECCV}, 2016.

\bibitem{aa}
Irwan Bello, Barret Zoph, Ashish Vaswani, Jonathon Shlens, and Quoc~V Le.
\newblock Attention augmented convolutional networks.
\newblock In {\em ICCV}, 2019.

\bibitem{mmdet}
Kai Chen, Jiaqi Wang, Jiangmiao Pang, Yuhang Cao, Yu Xiong, Xiaoxiao Li,
  Shuyang Sun, Wansen Feng, Ziwei Liu, Jiarui Xu, Zheng Zhang, Dazhi Cheng,
  Chenchen Zhu, Tianheng Cheng, Qijie Zhao, Buyu Li, Xin Lu, Rui Zhu, Yue Wu,
  Jifeng Dai, Jingdong Wang, Jianping Shi, Wanli Ouyang, Chen~Change Loy, and
  Dahua Lin.
\newblock {MMDetection}: Open mmlab detection toolbox and benchmark.
\newblock {\em arXiv preprint arXiv:1906.07155}, 2019.

\bibitem{chen2017deeplab}
Liang-Chieh Chen, George Papandreou, Iasonas Kokkinos, Kevin Murphy, and Alan~L
  Yuille.
\newblock Deeplab: Semantic image segmentation with deep convolutional nets,
  atrous convolution, and fully connected crfs.
\newblock {\em IEEE Transactions on Pattern Analysis and Machine Intelligence},
  40(4):834--848, 2017.

\bibitem{dynamic_conv}
Yinpeng Chen, Xiyang Dai, Mengchen Liu, Dongdong Chen, Lu Yuan, and Zicheng
  Liu.
\newblock Dynamic convolution: Attention over convolution kernels.
\newblock In {\em CVPR}, 2020.

\bibitem{crossmodal3}
Yukyung Choi, Namil Kim, Soonmin Hwang, Kibaek Park, Jae~Shin Yoon, Kyounghwan
  An, and In~So Kweon.
\newblock Kaist multi-spectral day/night data set for autonomous and assisted
  driving.
\newblock {\em IEEE Transactions on Intelligent Transportation Systems}, 2018.

\bibitem{control_brain}
Maurizio Corbetta and Gordon~L Shulman.
\newblock Control of goal-directed and stimulus-driven attention in the brain.
\newblock {\em Nature Reviews Neuroscience}, 2002.

\bibitem{deformable_conv}
Jifeng Dai, Haozhi Qi, Yuwen Xiong, Yi Li, Guodong Zhang, Han Hu, and Yichen
  Wei.
\newblock Deformable convolutional networks.
\newblock In {\em ICCV}, 2017.

\bibitem{crossmodal2}
Pingyang Dai, Rongrong Ji, Haibin Wang, Qiong Wu, and Yuyu Huang.
\newblock Cross-modality person re-identification with generative adversarial
  training.
\newblock In {\em IJCAI}, 2018.

\bibitem{imagenet}
Jia Deng, Wei Dong, Richard Socher, Li-Jia Li, Kai Li, and Li Fei-Fei.
\newblock Imagenet: A large-scale hierarchical image database.
\newblock In {\em CVPR}, 2009.

\bibitem{gadde2016superpixel}
Raghudeep Gadde, Varun Jampani, Martin Kiefel, Daniel Kappler, and Peter~V
  Gehler.
\newblock Superpixel convolutional networks using bilateral inceptions.
\newblock In {\em ECCV}, 2016.

\bibitem{res2net}
Shanghua Gao, Ming-Ming Cheng, Kai Zhao, Xin-Yu Zhang, Ming-Hsuan Yang, and
  Philip~HS Torr.
\newblock Res2net: A new multi-scale backbone architecture.
\newblock {\em IEEE Transactions on Pattern Analysis and Machine Intelligence},
  2019.

\bibitem{ham}
Bumsub Ham, Minsu Cho, and Jean Ponce.
\newblock Robust image filtering using joint static and dynamic guidance.
\newblock In {\em CVPR}, 2015.

\bibitem{gf}
Kaiming He, Jian Sun, and Xiaoou Tang.
\newblock Guided image filtering.
\newblock In {\em ECCV}, 2010.

\bibitem{resnet}
Kaiming He, Xiangyu Zhang, Shaoqing Ren, and Jian Sun.
\newblock Deep residual learning for image recognition.
\newblock In {\em CVPR}, 2016.

\bibitem{mobile_v3}
Andrew Howard, Mark Sandler, Grace Chu, Liang-Chieh Chen, Bo Chen, Mingxing
  Tan, Weijun Wang, Yukun Zhu, Ruoming Pang, Vijay Vasudevan, et~al.
\newblock Searching for mobilenetv3.
\newblock In {\em ICCV}, 2019.

\bibitem{mobile_v1}
Andrew~G Howard, Menglong Zhu, Bo Chen, Dmitry Kalenichenko, Weijun Wang,
  Tobias Weyand, Marco Andreetto, and Hartwig Adam.
\newblock Mobilenets: Efficient convolutional neural networks for mobile vision
  applications.
\newblock {\em arXiv preprint arXiv:1704.04861}, 2017.

\bibitem{senet}
Jie Hu, Li Shen, and Gang Sun.
\newblock Squeeze-and-excitation networks.
\newblock In {\em CVPR}, 2018.

\bibitem{dmsg}
Tak-Wai Hui, Chen~Change Loy, and Xiaoou Tang.
\newblock Depth map super-resolution by deep multi-scale guidance.
\newblock In {\em ECCV}, 2016.

\bibitem{batch_norm}
Sergey Ioffe and Christian Szegedy.
\newblock Batch normalization: Accelerating deep network training by reducing
  internal covariate shift.
\newblock {\em ICML}, 2015.

\bibitem{saliency_based}
Laurent Itti, Christof Koch, and Ernst Niebur.
\newblock A model of saliency-based visual attention for rapid scene analysis.
\newblock {\em IEEE Transactions on Pattern Analysis and Machine Intelligence},
  1998.

\bibitem{jampani2016learning}
Varun Jampani, Martin Kiefel, and Peter~V Gehler.
\newblock Learning sparse high dimensional filters: Image filtering, dense crfs
  and bilateral neural networks.
\newblock In {\em CVPR}, pages 4452--4461, 2016.

\bibitem{dynamic_filter}
Xu Jia, Bert De~Brabandere, Tinne Tuytelaars, and Luc~V Gool.
\newblock Dynamic filter networks.
\newblock In {\em NeurIPS}, 2016.

\bibitem{adam}
Diederik~P Kingma and Jimmy Ba.
\newblock Adam: A method for stochastic optimization.
\newblock {\em ICLR}, 2015.

\bibitem{jbu}
Johannes Kopf, Michael~F Cohen, Dani Lischinski, and Matt Uyttendaele.
\newblock Joint bilateral upsampling.
\newblock {\em ACM Transactions on Graphics}, 2007.

\bibitem{alexnet}
Alex Krizhevsky, Ilya Sutskever, and Geoffrey~E Hinton.
\newblock Imagenet classification with deep convolutional neural networks.
\newblock In {\em NeurIPS}, 2012.

\bibitem{psconv}
Duo Li, Anbang Yao, and Qifeng Chen.
\newblock Psconv: Squeezing feature pyramid into one compact poly-scale
  convolutional layer.
\newblock In {\em ECCV}, 2020.

\bibitem{djf}
Yijun Li, Jia-Bin Huang, Narendra Ahuja, and Ming-Hsuan Yang.
\newblock Deep joint image filtering.
\newblock In {\em ECCV}, 2016.

\bibitem{djf+}
Yijun Li, Jia-Bin Huang, Narendra Ahuja, and Ming-Hsuan Yang.
\newblock Joint image filtering with deep convolutional networks.
\newblock {\em IEEE Transactions on Pattern Analysis and Machine Intelligence},
  2019.

\bibitem{fpn}
Tsung-Yi Lin, Piotr Doll{\'a}r, Ross Girshick, Kaiming He, Bharath Hariharan,
  and Serge Belongie.
\newblock Feature pyramid networks for object detection.
\newblock In {\em CVPR}, 2017.

\bibitem{coco}
Tsung-Yi Lin, Michael Maire, Serge Belongie, James Hays, Pietro Perona, Deva
  Ramanan, Piotr Doll{\'a}r, and C~Lawrence Zitnick.
\newblock Microsoft coco: Common objects in context.
\newblock In {\em ECCV}, 2014.

\bibitem{texture}
Feng Liu and Michael Gleicher.
\newblock Texture-consistent shadow removal.
\newblock In {\em ECCV}, 2008.

\bibitem{darts}
Hanxiao Liu, Karen Simonyan, and Yiming Yang.
\newblock Darts: Differentiable architecture search.
\newblock {\em ICLR}, 2019.

\bibitem{weightnet}
Ningning Ma, Xiangyu Zhang, Jiawei Huang, and Jian Sun.
\newblock Weightnet: Revisiting the design space of weight networks.
\newblock In {\em ECCV}, 2020.

\bibitem{deconvolution}
Hyeonwoo Noh, Seunghoon Hong, and Bohyung Han.
\newblock Learning deconvolution network for semantic segmentation.
\newblock In {\em ICCV}, 2015.

\bibitem{bam}
Jongchan Park, Sanghyun Woo, Joon-Young Lee, and In~So Kweon.
\newblock Bam: Bottleneck attention module.
\newblock {\em BMVC}, 2018.

\bibitem{faster_rcnn}
Shaoqing Ren, Kaiming He, Ross Girshick, and Jian Sun.
\newblock Faster r-cnn: Towards real-time object detection with region proposal
  networks.
\newblock In {\em NeurIPS}, 2015.

\bibitem{dynamic_repr}
Ronald~A. Rensink.
\newblock The dynamic representation of scenes.
\newblock {\em Visual Cognition}, 2000.

\bibitem{mobile_v2}
Mark Sandler, Andrew Howard, Menglong Zhu, Andrey Zhmoginov, and Liang-Chieh
  Chen.
\newblock Mobilenetv2: Inverted residuals and linear bottlenecks.
\newblock In {\em CVPR}, 2018.

\bibitem{pixel_shuffle}
Wenzhe Shi, Jose Caballero, Ferenc Husz{\'a}r, Johannes Totz, Andrew~P Aitken,
  Rob Bishop, Daniel Rueckert, and Zehan Wang.
\newblock Real-time single image and video super-resolution using an efficient
  sub-pixel convolutional neural network.
\newblock In {\em CVPR}, 2016.

\bibitem{nyu}
Nathan Silberman, Derek Hoiem, Pushmeet Kohli, and Rob Fergus.
\newblock Indoor segmentation and support inference from rgbd images.
\newblock In {\em ECCV}, 2012.

\bibitem{pac}
Hang Su, Varun Jampani, Deqing Sun, Orazio Gallo, Erik Learned-Miller, and Jan
  Kautz.
\newblock Pixel-adaptive convolutional neural networks.
\newblock In {\em CVPR}, 2019.

\bibitem{dau}
Domen Tabernik, Matej Kristan, and Ale{\v{s}} Leonardis.
\newblock Spatially-adaptive filter units for compact and efficient deep neural
  networks.
\newblock {\em International Journal of Computer Vision}, 2020.

\bibitem{mnasnet}
Mingxing Tan, Bo Chen, Ruoming Pang, Vijay Vasudevan, Mark Sandler, Andrew
  Howard, and Quoc~V Le.
\newblock Mnasnet: Platform-aware neural architecture search for mobile.
\newblock In {\em CVPR}, 2019.

\bibitem{cond_seg}
Zhi Tian, Chunhua Shen, and Hao Chen.
\newblock Conditional convolutions for instance segmentation.
\newblock In {\em ECCV}, 2020.

\bibitem{vsgnet}
Oytun Ulutan, ASM Iftekhar, and Bangalore~S Manjunath.
\newblock Vsgnet: Spatial attention network for detecting human object
  interactions using graph convolutions.
\newblock In {\em CVPR}, 2020.

\bibitem{res_att_net}
Fei Wang, Mengqing Jiang, Chen Qian, Shuo Yang, Cheng Li, Honggang Zhang,
  Xiaogang Wang, and Xiaoou Tang.
\newblock Residual attention network for image classification.
\newblock In {\em CVPR}, 2017.

\bibitem{carafe}
Jiaqi Wang, Kai Chen, Rui Xu, Ziwei Liu, Chen~Change Loy, and Dahua Lin.
\newblock Carafe: Content-aware reassembly of features.
\newblock In {\em ICCV}, 2019.

\bibitem{solov2}
Xinlong Wang, Rufeng Zhang, Tao Kong, Lei Li, and Chunhua Shen.
\newblock Solov2: Dynamic, faster and stronger.
\newblock {\em arXiv preprint arXiv:2003.10152}, 2020.

\bibitem{cbam}
Sanghyun Woo, Jongchan Park, Joon-Young Lee, and In So~Kweon.
\newblock Cbam: Convolutional block attention module.
\newblock In {\em ECCV}, 2018.

\bibitem{crossmodal1}
Ancong Wu, Wei-Shi Zheng, Hong-Xing Yu, Shaogang Gong, and Jianhuang Lai.
\newblock Rgb-infrared cross-modality person re-identification.
\newblock In {\em ICCV}, 2017.

\bibitem{wu2018dynamic}
Jialin Wu, Dai Li, Yu Yang, Chandrajit Bajaj, and Xiangyang Ji.
\newblock Dynamic filtering with large sampling field for convnets.
\newblock In {\em ECCV}, 2018.

\bibitem{resnext}
Saining Xie, Ross Girshick, Piotr Doll{\'a}r, Zhuowen Tu, and Kaiming He.
\newblock Aggregated residual transformations for deep neural networks.
\newblock In {\em CVPR}, 2017.

\bibitem{smemvqa}
Huijuan Xu and Kate Saenko.
\newblock Ask, attend and answer: Exploring question-guided spatial attention
  for visual question answering.
\newblock In {\em ECCV}, 2016.

\bibitem{condconv}
Brandon Yang, Gabriel Bender, Quoc~V Le, and Jiquan Ngiam.
\newblock Condconv: Conditionally parameterized convolutions for efficient
  inference.
\newblock In {\em NeurIPS}, 2019.

\bibitem{dilated}
Fisher Yu and Vladlen Koltun.
\newblock Multi-scale context aggregation by dilated convolutions.
\newblock {\em ICLR}, 2016.

\bibitem{adaptive_filter}
Julio Zamora~Esquivel, Adan Cruz~Vargas, Paulo Lopez~Meyer, and Omesh Tickoo.
\newblock Adaptive convolutional kernels.
\newblock In {\em ICCV Workshops}, 2019.

\bibitem{scale_adapt_conv}
Rui Zhang, Sheng Tang, Yongdong Zhang, Jintao Li, and Shuicheng Yan.
\newblock Scale-adaptive convolutions for scene parsing.
\newblock In {\em ICCV}, 2017.

\bibitem{dynet}
Yikang Zhang, Jian Zhang, Qiang Wang, and Zhao Zhong.
\newblock Dynet: Dynamic convolution for accelerating convolutional neural
  networks.
\newblock {\em arXiv preprint arXiv:2004.10694}, 2020.

\end{thebibliography}
